\documentclass[twoside]{article}
\usepackage{algorithm}
\usepackage{hyperref}
\usepackage{algpseudocode}
\usepackage{subfig}
\usepackage{aistats2023}
\usepackage{graphicx}
\usepackage[T1]{fontenc}
\usepackage[round]{natbib}

\begin{document}

%

%

\twocolumn[

\aistatstitle{On the optimization and pruning for Bayesian deep learning}
\begin{center}
\text{Xiongwen Ke and Yanan Fan}\\
\text{School of mathematics and statistics, UNSW}
\end{center}
]

\begin{abstract}
The goal of Bayesian deep learning is to provide uncertainty quantification via the posterior distribution. However, exact inference over the weight space is computationally intractable due to the ultra-high dimensions of the neural network. Variational inference (VI) is a promising approach, but naive application on weight space does not scale well and often underperform on predictive accuracy. In this paper, we propose a new adaptive variational Bayesian algorithm to train neural networks on weight space that achieves high predictive accuracy. By showing that there is an equivalence to Stochastic Gradient Hamiltonian Monte Carlo(SGHMC) with preconditioning matrix, we then propose an MCMC within EM algorithm, which incorporates the spike-and-slab prior to capture the sparsity of the neural network.
The EM-MCMC algorithm allows us to perform optimization and model pruning within one-shot. We evaluate our methods on CIFAR-10, CIFAR-100 and ImageNet datasets, and demonstrate that our dense model can reach the state-of-the-art performance and our sparse model perform very well compared to previously proposed pruning schemes. 
\end{abstract}

\section{INTRODUCTION}

Bayesian inference \citep{bishop2006pattern} provide an elegant way to capture uncertainty via the posterior distribution over model parameters. Unfortunately, posterior inference is intractable in any reasonably sized neural network models. 

Works focussing on scalable inference for Bayesian deep learning over the last decade can be separated by two streams. One is a deterministic approximation approach such as variational inference \citep{graves2011practical,blundell2015weight}, dropout \citep{gal2016dropout}, Laplace approximation \citep{ritter2018scalable}, or expectation propagation\citep{hernandez2015probabilistic}. The other stream involve sampling approaches such as MCMC using stochastic gradient Langevin dynamics (SGLM) \citep{welling2011bayesian,chen2014stochastic}.

Prior to 2019, Bayesian neural networks (BNN) generally struggle with predictive accuracy and computational efficiency. Recently, a lot of advances have been made in both directions of research. In deterministic approach, several authors began to consider using dimension reduction techniques such as subspace inference \citep{maddox2019simple}, rank-1 parameterization\citep{dusenberry2020efficient}, subnetwork inference\citep{daxberger2021bayesian} and node-space inference\citep{trinh2022tackling}. In the sampling approach, \citep{zhang2019cyclical} propose to use cycles of learning rates with a high-to-low step size schedule. A large step size in the early stage of the cycle results in aggressive exploration in the parameter space; as the step size decreases, the algorithm begins to collect sample around the local mode.

Apart from the progress within these two streams, \cite{wilson2020bayesian} show that deep ensembles \citep{lakshminarayanan2017simple} can be interpreted as an approximate approach to posterior predictive distribution.  They combine multiple independently trained SWAG (Gaussian stochastic weight averaging) approximations \citep{maddox2019simple,izmailov2018averaging}  to create a mixture of Gaussians approximation to the posterior.
However, performing variational inference directly on weight space still produces poor predictive accuracy and struggles with computational efficiency. Even a simple mean-field variational inference will involve double the number of parameters of the  neural network (i.e., mean and variance for each weight), which incurs a extra GPU memory requirement and  2-5 times of the runtime of the baseline neural network \citep{osawa2019practical}. 

In this paper, we develop an adaptive optimization algorithm for Gaussian Mean-field variational Bayesian inference that can achieve state-of-the-art predictive accuracy. We further show that when the learning rate is small and the update of the posterior variance has been frozen, the algorithm is equivalent to the SGHMC (Stochastic Gradient Hamiltonian Monte Carlo) with preconditioning matrix. Therefore, if we exploit the closed form expression of the gradient for the posterior variances and only keep track of the weight generated by the algorithm(no need to register the mean and variance parameter in the code), we can achieve big savings on 
GPU memory and runtime costs. 

Based on the connection to SGHMC, we extend the EM Algorithm for Bayesian variable selection \citep{rovckova2014emvs,wang2016ensemble,rovckova2018particle} for linear models to neural networks by replacing the Gaussian prior in BNN with the spike-and-slab group Gaussian prior\citep{xu2015bayesian}. Our method is an MCMC within EM algorithm, which will switch the weight decay factor between small and large based on the magnitude of each group during training. Since by construction, there are no exact zeros, we further find a simple pruning criterion to remove the weights permanently during training. A sparse model will be trained in one-shot without additional retrain. Our approach is more computationally efficient than those dynamic pruning strategies that allows regrow \citep{zhu2017prune,dettmers2019sparse,lin2020dynamic}. We will show that this aggressive approach has no performance loss. Our code is available at github: \url{https://github.com/z5041294/optimization-and-pruning-for-BNN}

\section{Optimization}
\subsection{Preliminaries on variational Bayesian neural network}
Given a dataset $\mathcal{D} = \left\{x_{i},y_{i}\right\}_{i=1}^{N}$, a Bayesian neural network (BNN) is defined in terms of a prior $p(\mathbf{w)}$ on the $p$-dimensional weights, as well as the neural network likelihood $p(\mathcal{D}|\mathbf{w})$. Variational Bayesian methods approximates the true posterior $p(\mathbf{w}|\mathcal{D})$ by minimising the KL divergence between the approximate distribution, $q_{\theta}(\mathbf{w})$, and the true posterior. This is shown to be equivalent to maximizing the evidence lower bound (ELBO):
\begin{equation}
\mathcal{L}[\theta]=\mathbf{E}_{q_{\theta}}[\log p(\mathcal{D} \mid \mathbf{w})]-\mathrm{D}_{\mathrm{KL}}(q_{\theta}(\mathbf{w}) \| p(\mathbf{w}))
\end{equation}
where we consider a Bayesian neural net with Gaussian prior $p(\mathbf{w})\sim N_{p}(0, \boldsymbol{\Sigma}_{0})$ and a Gaussian approximate posterior $q_{\theta}(\mathbf{w})\sim N_{p}(\boldsymbol{\mu}, \boldsymbol{\Sigma})$ where $\theta=(\boldsymbol{\mu}, \boldsymbol{\Sigma})$. 
To make it scale to large sized models, we assume both the approximate posterior and prior weights are independent, such that $p(w_{j}) \sim N(0,\delta^{-1})$ and $q_{\theta_{j}}(w_{j}) \sim N(\mu_{j},\sigma_{j}^{2})$. The gradient of ELBO with respect to $\boldsymbol{\mu}$ and $\boldsymbol{\Sigma}$ is
{\small
\begin{equation}
\begin{aligned}
\nabla_{\boldsymbol{\mu}}\mathcal{L} & =\mathbf{E}_{N_{p}(\boldsymbol{\mu}, \boldsymbol{\Sigma})}\left[\nabla_{\mathbf{w}} \log p(\mathcal{D} \mid \mathbf{w})\right]-\boldsymbol{\Sigma}_{0}^{-1}\boldsymbol{\mu} \approx -\frac{1}{S}\sum_{i=1}^{S}\mathbf{g}_{i}-\delta\mu \\
\nabla_{\boldsymbol{\Sigma}}\mathcal{L} & =\frac{1}{2} \mathbf{E}_{N_{p}(\boldsymbol{\mu}, \boldsymbol{\Sigma})}\left[\nabla_{\mathbf{w}}^{2} \log p(\mathcal{D} \mid \mathbf{w})\right]+\frac{1}{2} \boldsymbol{\Sigma}^{-1}-\frac{1}{2}\boldsymbol{\Sigma}_{0}^{-1}\\
& \approx -\frac{1}{2S}\sum_{i=1}^{S}\mathbf{g}_{i}^{2}+\frac{1}{2}\mathbf{diag}(\sigma_{j}^{-2})-\frac{1}{2}\delta
\end{aligned}
\end{equation}
}%
where $\mathbf{g}_{i}=-\nabla_{\mathbf{w}} \log p(\mathcal{D} \mid \mathbf{w_{i}})$ and $\mathbf{w}_{i}\sim \prod_{j=1}^{p}N(\mu_{j},\sigma_{j}^{2})$ are Monte Carlo samples. In addition, we also have the second order derivative of $\mu$
$$
{\small
\begin{aligned}
\nabla_{\boldsymbol{\mu}}^{2}\mathcal{L} & =-\mathbf{E}_{N_{p}(\boldsymbol{\mu}, \boldsymbol{\Sigma})}\left[\nabla^{2}_{\mathbf{w}} \log p(\mathcal{D} \mid \mathbf{w})\right]+ \boldsymbol{\Sigma}_{0}^{-1} \approx  \frac{1}{S}\sum_{i=1}^{S}\mathbf{g}_{i}^{2}+\delta\\
\end{aligned}
}%
$$
Using Gaussian back-propagation and reparameterization trick, an alternative MC approximation \citep{khan2018fast} can be used,  such that
$-\mathbf{E}_{N_{p}(\boldsymbol{\mu}, \boldsymbol{\Sigma})}\left[\nabla^{2}_{\mathbf{w}} \log p(\mathcal{D} \mid \mathbf{w})\right] \approx \frac{1}{S}\sum_{i=1}^{S}\mathbf{g}_{i}\odot \frac{\mathbf{\epsilon}_{i}}{\mathbf{\sigma}}$, where $\mathbf{\epsilon}_i\sim N(0, \mathbf{1}_p)$. In practice, to reduce the runtime complexity, we often use $S=1$ as long as the batch size in one iteration is not too small.

\subsection{Bayesian versions of adaptive algorithm}
In Gaussian mean-field  variational Bayesian inference, the natural-gradient descent\citep{khan2018fast,zhang2018noisy,osawa2019practical} optimizes the ELBO bound by updating
$$
{\small
\begin{aligned}
\boldsymbol{\mu}_{t+1} & =\underset{\boldsymbol{\mu} \in \mathbf{R}^{p}}{\operatorname{argmin}}\left\{\langle\nabla_{\boldsymbol{\mu}} \mathcal{L}, \boldsymbol{\mu}\rangle+\frac{1}{2 l_{t}}(\boldsymbol{\mu}-\boldsymbol{\mu}_{t})^{T}\mathbf{diag}(\mathbf{\sigma}^{-2})^{\alpha}(\boldsymbol{\mu}-\boldsymbol{\mu}_{t})^{T}\right\}\\
& = \boldsymbol{\mu}_{t}-l_{t}(\boldsymbol{\sigma}_{t}^{2})^{\alpha} \odot \nabla_{\boldsymbol{\mu}_{t}}\mathcal{L} 
\end{aligned}
}%
$$
where $l_{t}$ is a learning rate, $\alpha \in \left\{\frac{1}{2},1\right\}$
and $\frac{1}{\boldsymbol{\sigma}_{t}^{2}}$ is updated with momentum $\frac{1}{\boldsymbol{\sigma}_{t}^{2}}= \frac{1-\lambda\gamma}{\boldsymbol{\sigma}_{t-1}^{2}}+\gamma([\boldsymbol{g}_{t}\odot \boldsymbol{g}_{t}]+\frac{\delta}{N})$ ($0<\gamma<1$ and $\lambda$ is another learning rate). When $\alpha=\frac{1}{2}$, this is similar to the Adam \citep{kingma2014adam} algorithm and when $\alpha=1$, the algorithm is very close to second-order optimization with the Hessian matrix given by $\boldsymbol{\Sigma}^{-1}$ when $\nabla_{\boldsymbol{\Sigma}}\mathcal{L}=0$ in equation (2).

There are two concerns for this version of Bayesian adaptive algorithm:  First, similar to Adam, the variance of adaptive learning rate is problematically large in the early stages of the optimization \citep{liu2019variance}. A warm-up stage is recommended for traditional Adam. As an example,  consider a ReLu neural network with a single hidden layer and binary cross entropy loss, then if a normal initialization of the weight with mean zero has been used, then the variance of the adaptive learning will diverge at the beginning of the training period (See detailed discussion in the Appendix A).
Second,  when the number of Monte Carlo samples for the gradient is $S=1$ and the learning rate is small, the injected Gaussian noise may dominate the gradient.  We see that  
$$
\begin{aligned}
\mathbf{w}_{t}-\mathbf{w}_{t-1}  & =\boldsymbol{\mu}_{t}-\boldsymbol{\mu}_{t-1}+(\boldsymbol{\sigma}_{t} \odot \boldsymbol{\epsilon}-\boldsymbol{\sigma}_{t-1}\odot \boldsymbol{\epsilon}')\\
& = -l_{t}(\boldsymbol{\sigma}_{t}^{2})^{\alpha}\odot \boldsymbol{m}_{t}+ \boldsymbol{\epsilon} \odot \sqrt{\boldsymbol{\sigma}_{t}^{2}+\boldsymbol{\sigma}_{t-1}^{2}}\\
\end{aligned}
$$
where $\boldsymbol{m}_{t}$ is the momentum of gradient. For $\alpha=\frac{1}{2}$, when the learning rate is small, the gradient may be erased by the injected noise. For $\alpha=1$, this could also happen when $\boldsymbol{\sigma}_{t}^{2} \ll 1$. Empirically, we observe that both $\mathbf{w} \ll 1$ and $\boldsymbol{\sigma}^{2} \ll 1$ in Bayesian deep learning. So a warm up strategy that starts with a very small learning rate may not work for variational BNN. On the contrary, when the learning rate is large, the size of the injected noise will be relatively small, the model could overfit. In summary, there is a delicate balance between the size of the gradient and the size of the injected noise. This argument will become more lucid when we show the connection between our algorithm and SGHMC.

\subsection{Constrained variational Adam}
To remedy the issue highlighted above, we reparameterize $\boldsymbol{\sigma}_{i}$ as a product of global and local parameters such that $\boldsymbol{\sigma}_{i}=\alpha \boldsymbol{\tau}_{i}$ where $\boldsymbol{\tau}_{i} \in (0,1)$ and $\alpha>0$.  We treat $\alpha$ as a hyper-parameter, which can be used in an annealing scheme.  Now, the variance of the adaptive learning rate is upper bounded by $\frac{\alpha^{2}}{12}$. The updated posterior mean is modified to
$$
{\small
\begin{aligned}
\boldsymbol{\mu}_{t+1} & =\underset{\boldsymbol{\mu} \in \mathbf{R}^{p}}{\operatorname{argmin}}\left\{\langle\nabla_{\boldsymbol{\mu}} \mathcal{L}, \boldsymbol{\mu}\rangle+\frac{1}{2 l_{t}}(\boldsymbol{\mu}-\boldsymbol{\mu}_{t})^{T}\mathbf{diag}(\boldsymbol{\tau}^{-1})(\boldsymbol{\mu}-\boldsymbol{\mu}_{t})^{T}\right\}\\
& = \boldsymbol{\mu}_{t}-l_{t}\boldsymbol{\tau}_{t} \odot \nabla_{\boldsymbol{\mu}_{t}}\mathcal{L} 
\end{aligned}
}%
$$
Since $\boldsymbol{\tau}_{i}\in(0,1)$ are the parameters we need to learn, the objective function for updating the posterior variance change to:
$$
\begin{aligned}
\boldsymbol{\tau}_{t+1} & =\underset{\boldsymbol{\tau} \in (0,1)^{p}}{\operatorname{argmin}}\left\{\langle\nabla_{\boldsymbol{\tau}} \mathcal{L}, \boldsymbol{\tau}\rangle+\frac{1}{2 \eta}\|\boldsymbol{\tau}-\boldsymbol{\tau}_{t}\|^{2}\right\}\\
           & = \underset{\boldsymbol{\tau} \in (0,1)^{p}}{\operatorname{argmin}}\|(\boldsymbol{\tau}_{t}-\eta\nabla_{\boldsymbol{\tau}} \mathcal{L})-\tau\|^{2}\\
           & = \pi_{(0,1)^{P}}(\boldsymbol{\tau}_{t}-\eta\nabla_{\boldsymbol{\tau}} \mathcal{L})
\end{aligned}
$$
where $\eta$ is a learning rate, $\pi_{(0,1)^{P}}$ denotes the Euclidean projection onto $(0,1)^{P}$.  Now we replace the Euclidean distance of  $\frac{1}{2}\|\boldsymbol{\tau}_{t+1}-\boldsymbol{\tau}_{t}\|^{2}$ to Bregman distance $D_{G}(\boldsymbol{\tau}_{t+1}, \boldsymbol{\tau}_{t})$, where $D_{G}(\cdot,\cdot)$ is generated by a strictly convex twice-differentiable function $G(\cdot)$
$$D_{G}(\boldsymbol{\tau}, \boldsymbol{\tau}^{\prime}):=G(\boldsymbol{\tau})-G(\boldsymbol{\tau}^{\prime})-\langle\nabla G(\boldsymbol{\tau}^{\prime}), \boldsymbol{\tau}-\boldsymbol{\tau}^{\prime}\rangle.$$
If $G=\frac{1}{2}\|\cdot\|^{2}$, we recover the squared Euclidean distance. This modification leads to a more general version of gradient descent algorithm which is known as mirror descent.

The idea is that we want to find a distance function which can better reflect the geometry of $(0,1)^{p}$. Now our objective function can be written as a more general form:
\begin{equation}
\boldsymbol{\tau}_{t+1}  =\underset{\boldsymbol{\tau} \in (0,1)^{p}}{\operatorname{argmin}}\left\{\langle\nabla_{\boldsymbol{\tau}} \mathcal{L}, \boldsymbol{\tau}\rangle+\frac{1}{\eta}D_{G}(\boldsymbol{\tau}, \boldsymbol{\tau}_{t})\right\}
\end{equation}
The choice of $G(\cdot)$ will define the geometry and the distance metric of the primal space and dual space. Values in $(0,1)$ may be interpreted as a probability. We define the distance $D_{G}(\boldsymbol{\tau}, \boldsymbol{\tau}_{t})$ as negative binary cross entropy loss:
$$G(\tau)=\tau\log \tau+(1-\tau)\log(1-\tau)$$
Then parameter in the dual space is defined as
\begin{equation}\label{dual}
\rho =\nabla_{\tau} G(\tau)=g(\tau)=\log(\tau)-\log(1-\tau)
\end{equation}
with the inverse function from dual space to the primal space, which is the logistic sigmoid function
\begin{equation}\label{inverse_function}
\tau=\nabla_{\rho} H(\rho)=h(\rho)=g^{-1}(\rho)=\frac{1}{1+\exp(-\rho)}
\end{equation}
The mirror descent allows us to perform gradient descent in the dual space, which is unconstrained in our case, as shown in equation (\ref{dual}), and finally move back to the primal space by equation (\ref{inverse_function}). Algorithm 1 provides the psudo-code for the constrained variational Adam. 
(Appendix B provides a quick introduction of mirror descent and the comparison with the reparameterization trick).
\begin{algorithm}
\caption{Constrained variational Adam}
\begin{algorithmic}
\State \textbf{Input:} $\beta=(\beta_1,\beta_2)$,$\delta \text{(weight decay)}$\\
\quad \quad \,\,$N\text{(training size)}$, $B\text{(batch size)}$ 
\State\textbf{Initialization:} $\boldsymbol{m}$, $\boldsymbol{\rho}$, $\boldsymbol{\mu}$, $\boldsymbol{\tau}$
\For{$t=1 \: \cdots \: $}
\State $l \gets s_{1}(t)$ \Comment{learning rate schedule}
\State $\eta \gets s_{2}(t)$ \Comment{learning rate schedule}
\State $\alpha \gets s_{3}(t)$\Comment{annealing schedule}
\State $\textbf{Sample} \: \boldsymbol{\epsilon} \sim N(0,\mathbf{1}_{p})$
\State $\mathbf{w}   \gets  \boldsymbol{\mu}+\alpha \boldsymbol{\tau}\odot \boldsymbol{\epsilon}$
\State $\boldsymbol{g} \gets -\frac{1}{B}\sum_{i \in \mathcal{B}}\nabla_{\mathbf{w}} \log(\mathcal{D}_{i}|\mathbf{w})+\frac{\delta}{N} \boldsymbol{\mu}$
\State  $\boldsymbol{m}   \gets  \beta_1 \boldsymbol{m} + \beta_2 \boldsymbol{g}$
\State $\boldsymbol{\rho}  \gets   \boldsymbol{\rho}+ \left(\frac{\eta}{\tau}-\eta\alpha^{2}\delta \boldsymbol{\tau}\right)- \eta\alpha\boldsymbol{\epsilon} \odot \boldsymbol{g}$
\State $\boldsymbol{\tau} \gets   1/(1+e^{-\boldsymbol{\rho}})$
\State $\boldsymbol{\mu} \gets \boldsymbol{\mu} - l \boldsymbol{\tau} \odot \boldsymbol{m}$
\EndFor
\end{algorithmic}
\end{algorithm}

\subsection{Connection to SGHMC}
If we set the global parameter of posterior standard deviation $\alpha=kl_{t}^{3/4}$ for some positive $k$ and $\beta_{1}=1-hl_{t}^{1/2}$ for some positive $h$, then based on the following two assumptions: 
\begin{itemize}
    \item the learning rate is small and $l_{t} \approx l_{t-1}$
    \item the update of $\tau$ is frozen or it has already converged
\end{itemize}
then the constrained variational Adam algorithm is equivalent to SGHMC\citep{chen2014stochastic} 
\begin{equation}\label{SGHMC}
\begin{aligned}
d\mathbf{w}_{t} & = D_{\boldsymbol{\tau}}\boldsymbol{r}_{t}dt\\
d\boldsymbol{r}_{t} & =-\beta_{2}\boldsymbol{g}_{t}dt-h\boldsymbol{r}_{t}dt+\sqrt{2+2\beta_{1}^{2}}k d\boldsymbol{B}_{t}
\end{aligned}
\end{equation}
where $D_{\boldsymbol{\tau}}=\mathbf{Diag}(\boldsymbol{\tau})$ is a preconditioning matrix and $d\boldsymbol{B}_{t}$ is Brownian motion. In Appendix C, we show this relationship by setting $\sqrt{l_{t}}=\Delta t$. The above dynamic produces the distribution proportional to $\exp\left(-\frac{\beta_{2} L(\mathbf{w})}{k^{2}}\right)$ where $L(\mathbf{w})=-\frac{1}{N}\sum_{i=1}^{N}\log p(y_{i}|x_{i},\mathbf{w})-\frac{1}{N}\log p(\mathbf{w})$. By setting $\beta_{2}=1$ and $k=\frac{1}{\sqrt{N}}$, we recover the posterior of the weight $p(\mathbf{w}|\mathcal{D})$. 

Based on this connection, it is natural to consider using the cyclical learning rate \citep{zhang2019cyclical} with high-to-low learning rate in each cycle. A high learning rate at the beginning can quickly find a local mode and a small learning rate at the end ensures an accurate simulation around the mode.

From Gaussian variational inference perspective, $\alpha$ is the annealing parameter, which controls the temperature of the approximated posterior. Unlike the original Bayesian backpropagation \citep{blundell2015weight}, we actually introduce an extra inductive bias into the approximate posterior by only training the local variance parameter $\tau$ and defining the global variance parameter $\alpha$. We believe that, in high dimension, the magnitude of posterior variance for individual weight should be very small.

When the size of the model is large, to improve the predictive performance, we use the cold posterior \citep{heek2019bayesian,zhang2019cyclical,wenzel2020good} by setting $k=\frac{1}{N}$ rather than $\frac{1}{\sqrt{N}}$. By defining $\boldsymbol{v}_{t}=\boldsymbol{r}_{t-1}\Delta t$, \citep{chen2014stochastic}, we derive
the numerical scheme shown in algorithm 2. \\
\textbf{Remark}: The implementation of constrained variational Adam(CV-Adam) algorithm is always based on Algorithm 2. In addition, we found that it is not necessary to freeze the update of $\tau$ as it converges well in practice. 
\begin{algorithm}
\centering
\caption{SGHMC}
\begin{algorithmic}
\State \textbf{Input} $\beta=(\beta_{1},\beta_2)$, $\delta \text{(weight decay)}$\\
\quad \quad \,\,$N\text{(training size)}$, $B\text{(batch size)}$
\State\textbf{Initialization}  $\boldsymbol{\rho}$,$\mathbf{w}$,$\boldsymbol{v}$
\For{$t=1 \: \cdots \: $}
\State $\Delta t \gets s_{1}(t)$  \Comment{learning rate schedule}
\State $\eta \gets s_{2}(t)$      \Comment{learning rate schedule}
\State $k \gets s_{3}(t)$          \Comment{annealing schedule}  
\State $\alpha \gets k*\Delta t^{1.5}$ 
\State $\boldsymbol{g} \gets -\frac{1}{B}\sum_{i \in \mathcal{B}}\nabla_{\mathbf{w}}\log(\mathcal{D}_{i}|\mathbf{w})+\frac{\delta}{N} \mathbf{w}$
\State $\boldsymbol{\rho}  \gets   \boldsymbol{\rho}+ \left(\frac{\eta}{\boldsymbol{\tau}}-\eta\alpha^{2}\delta \boldsymbol{\tau}\right)- \eta\alpha\boldsymbol{\epsilon} \odot \boldsymbol{g}$
\State $\boldsymbol{\epsilon} \gets N(0,\mathbf{1}_{p})$
\State $\boldsymbol{\tau} \gets   1/(1+e^{-\boldsymbol{\rho}})$
\State $\boldsymbol{v} \gets \small{\beta_{1}\boldsymbol{v}-\beta_{2}\boldsymbol{g}(\Delta t)^{2}+\sqrt{2+2\beta_{1}^{2}}k(\Delta t)^{1.5}\boldsymbol{\epsilon}}$
\State $\mathbf{w} \gets \mathbf{w} - \boldsymbol{\tau} \odot \boldsymbol{v}$
\EndFor
\end{algorithmic}
\end{algorithm}

\section{Pruning}
In this section, we describe how we construct an efficient pruning scheme to obtain a sparse network for Bayesian deep learning. 

\subsection{Spike-and-Slab prior}

We apply the group spike and slab prior to the weight parameters in the neural network. For simplicity, we ignore the subscript for the layers and consider the weights between any two layers. For convolutional layer,  let $w_{ij}=(w_{ij1},...,w_{ijK^{2}})$ denote the group of $K^{2}$ parameters from the $i$th input channel to $j$th output channel, and $K\times K$ is the size of the kernel. Then the  prior for $w_{ijk}$ conditioned on a binary inclusion parameter $\gamma_{ij}$ for the entire group of weights, follows an independent normal distribution such that:
\begin{equation}\label{SPL_prior}
\pi\left(w_{ijk} \mid  \gamma_{ij} \right)= \begin{cases}\mathrm{N}\left(0,  \delta_{1}^{-1}\right) & \text { if } \gamma_{ij}=1 \\\mathrm{N}\left(0,  \delta_{0}^{-1}\right) & \text { otherwise, } \end{cases}
\end{equation}
where  $\pi(\gamma_{ij})$ are i.i.d binary distribution such that 
$$
\pi\left(\gamma_{ij}\right)= \begin{cases}1-p_{ij} & \text { if } \gamma_{ij}=1 \\p_{ij} & \text { if } \gamma_{ij}=0.\end{cases}
$$
The basic assumption is that a priori, within the same kernel all the weight parameters have the same distribution. For a fully connected layer, we can also make groupings based on whether they share the same input or output unit, for instance, if a group is made based on the input unit, then this is similar to variable/feature selection. Alternatively, we can assign the spike-and-slab-Gaussian prior to all the weights individually, this is related to unstructured pruning. 

\subsection{The EM algorithm}

We now derive a conditional EM algorithm that returns a MAP estimator of $\gamma_{ij}$. With the likelihood from the neural network model and prior distributions specified above, we write the full posterior distribution as follows:
$$
\pi\left(\boldsymbol{\gamma}, \mathbf{w} \mid \mathbf{y}\right) \propto \underset{ijk}{\prod} p\left(\mathbf{y} \mid w_{ij}\right)  \pi(w_{ijk}\mid \boldsymbol{\gamma}_{ij})  \pi(\boldsymbol{\gamma}_{ij} ).
$$
\textbf{E-Step}:The objective function $Q$ at the $t$th iteration in an EM algorithm is defined as
the integrated logarithm of the full posterior with respect to $w_{ij}$
$
\begin{aligned}
Q\left(\gamma_{ij} \mid \gamma_{ij}^{(t-1)}\right) & =\mathbf{E}_{\pi(\mathbf{w} \mid \gamma_{ij}^{(t-1)},\gamma_{-(ij)}, \mathbf{y})} \log \pi(\boldsymbol{\gamma}, \mathbf{w} \mid \mathbf{y})\\
& = \mathbf{E}_{\pi(\mathbf{w} \mid \gamma_{ij}^{(t-1)},\gamma_{-(ij)}, \mathbf{y})} \log \pi(y \mid w)\\
& -\frac{1}{2}E_{\pi(\mathbf{w} \mid \gamma_{ij}^{(t-1)},\gamma_{-(ij)}, \mathbf{y})}\sum_{k}[(\delta_{1}-\delta_{0})\gamma_{ij}+\delta_{0}]w_{ijk}^{2}\\
& +\frac{K^{2}}{2}\log[\delta_{0}+(\delta_{1}-\delta_{0})\gamma_{ij}]+\log\pi(\gamma_{ij})+C
\end{aligned}
$
\textbf{M-Step}: 
Set $\gamma_{ij}=0$,  if $Q\left(\gamma_{ij}=0 \mid\gamma_{ij}^{(t-1)}\right)\geq  Q\left(\gamma_{ij}=1 \mid \gamma_{ij}^{(t-1)}\right)$,that is,
$$
{\tiny
\begin{aligned}
\frac{E_{\pi(\mathbf{w} \mid \gamma_{ij}^{(t-1)},\gamma_{-(ij)}, \mathbf{y})}[\sum_{k}w_{ijk}^{2}]}{K^{2}}  \leq \frac{1}{\delta_{0}-\delta_{1}}\left[\log \frac{\delta_{0}}{\delta_{1}}+\frac{2}{K^{2}}\log\frac{p_{ij}}{1-p_{ij}}\right]=\lambda_{1}
\end{aligned}
}%
$$
and 1 otherwise. Since the term $\log\left(\frac{p_{ij}}{1-p_{ij}}\right) \in \mathbf{R}$ and $0<p_{ij}<1$ is the hyper-parameter,instead of turning $p_{ij}$, we can turn the  threshold $\lambda_{1}$ on the right hand side directly. 

For neural networks,  $\pi(\mathbf{w}|\boldsymbol{y},\boldsymbol{\gamma}) \propto \pi(\boldsymbol{y}|\mathbf{w})\pi(\mathbf{w}|\boldsymbol{\gamma})$ has no closed form,  an approximation can be obtained by using the samples produced by our constrained variational Adam algorithms:
$E_{\pi(\mathbf{w}|\gamma_{ij}^{(t-1)},\gamma_{-(ij)},y)}[\sum_{k}w_{ijk}^{2}] \approx \sum_{k}w_{ijk}^{2}$.

Our method can be viewed as a hybird MCMC or variational with EM algorithm, which will adaptively change the weight decay factor for each group based on the magnitude. A large weight decay factor will be assigned to the group of weights with small magnitude via $\frac{\delta_0}{N}$. This will further push them toward zeros. On the contrary, the group with strong signal will assign a small weight decay factor, via $\frac{\delta_1}{N}$.

\subsection{Dynamic forward pruning (DFP)}
Since our EM-MCMC algorithm produces no exact zeros, a hard threshold is required to prune the weight and maintain sparsity. Recently there are large number of dynamic pruning methods that maintain sparse weights through a prune-regrowth cycle during training \citep{zhu2017prune,bellec2017deep,dettmers2019sparse,mostafa2019parameter,lin2020dynamic,kusupati2020soft}. Some of these methods can be inserted into our framework directly. For example, we can incorporate dynamic pruning with feedback (DPF) \citep{lin2020dynamic}, which evaluates a stochastic gradient for the pruned model $\widetilde{\mathbf{w}}_t=\mathbf{m}_t \odot \mathbf{w}_t$ (where $\mathbf{m}_t$ is a mask matrix) and apply it to the (simultaneously maintained) dense model
\begin{equation}\label{DPF}
\mathbf{w}_{t+1}:=\mathbf{w}_t-\gamma_t \mathbf{g}\left(\mathbf{m}_t \odot \mathbf{w}_t\right)=\mathbf{w}_t-\gamma_t \mathbf{g}\left(\widetilde{\mathbf{w}}_t\right)
\end{equation}
and $\mathbf{g}\left(\mathbf{m}_t \odot \mathbf{w}_t\right)$ is the gradient evaluated at the sparse model. Although these methods are one-shot, some computational budget for weight regrow is still required. 

For linear models, it has been argued that forward or backward selection strategy may be trapped into a bad local mode \citep{hastie2009elements}, thus a variable selection algorithm allowing parameter regrow is preferred. 

\begin{figure}[t]
    \centering
    \subfloat[\centering Layer 1 Input channel  3]{{\includegraphics[width=2cm]{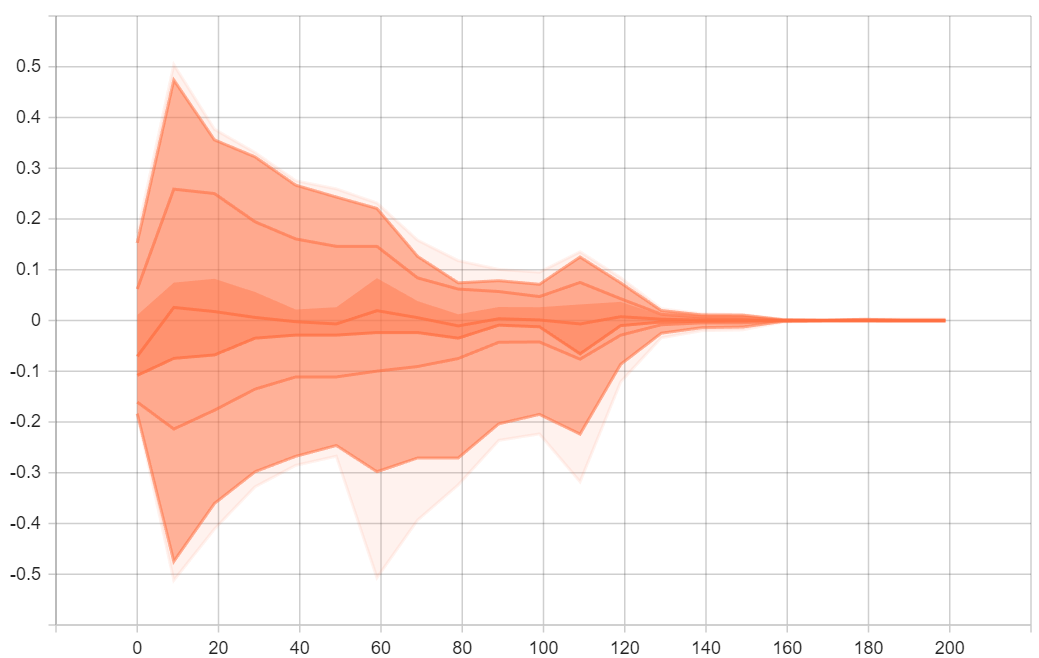} }}%
    \qquad
    \subfloat[\centering Layer 6 Input channel   20]{{\includegraphics[width=2cm]{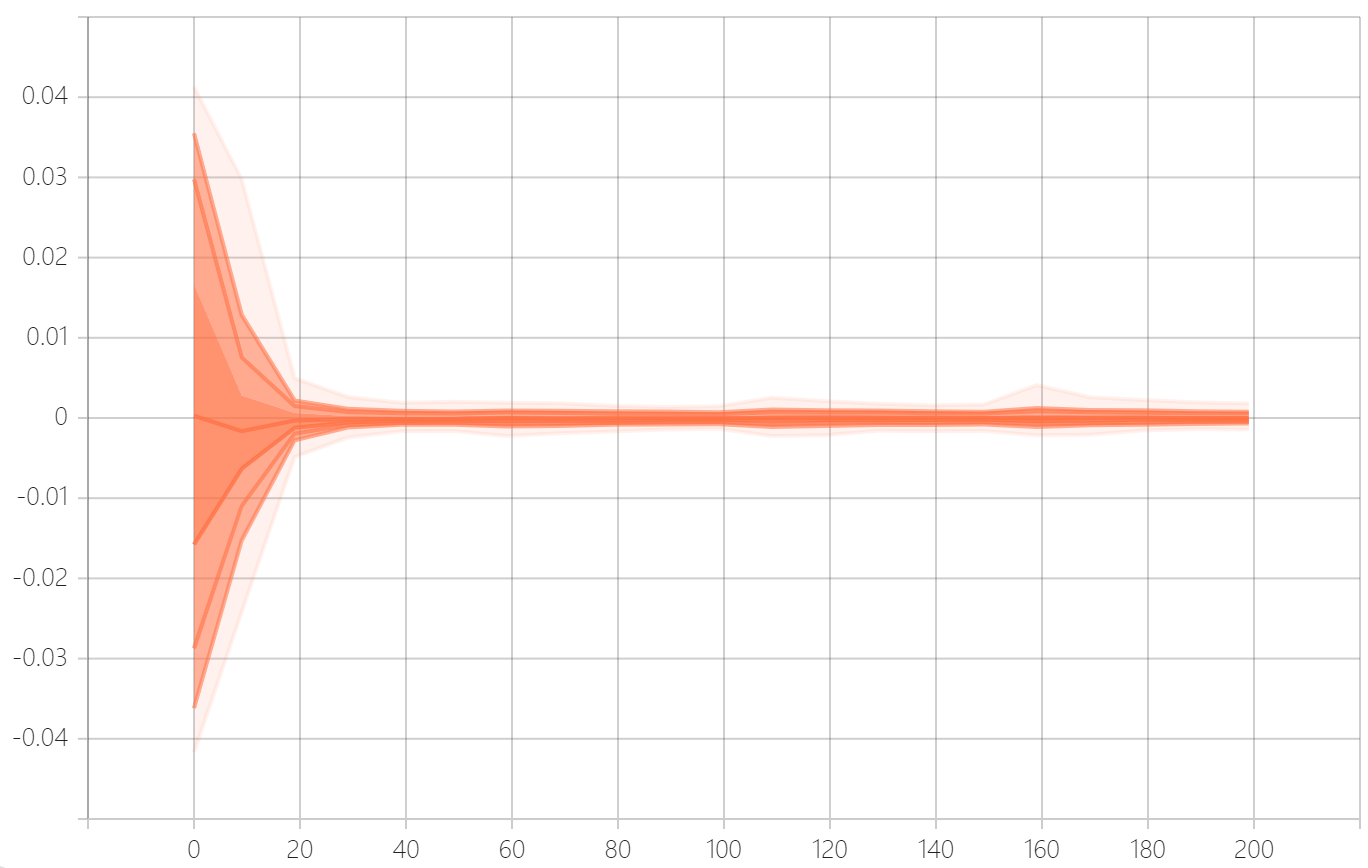} }}%
    \qquad
    \subfloat[\centering Layer 6 Output channel 14 ]{{\includegraphics[width=2cm]{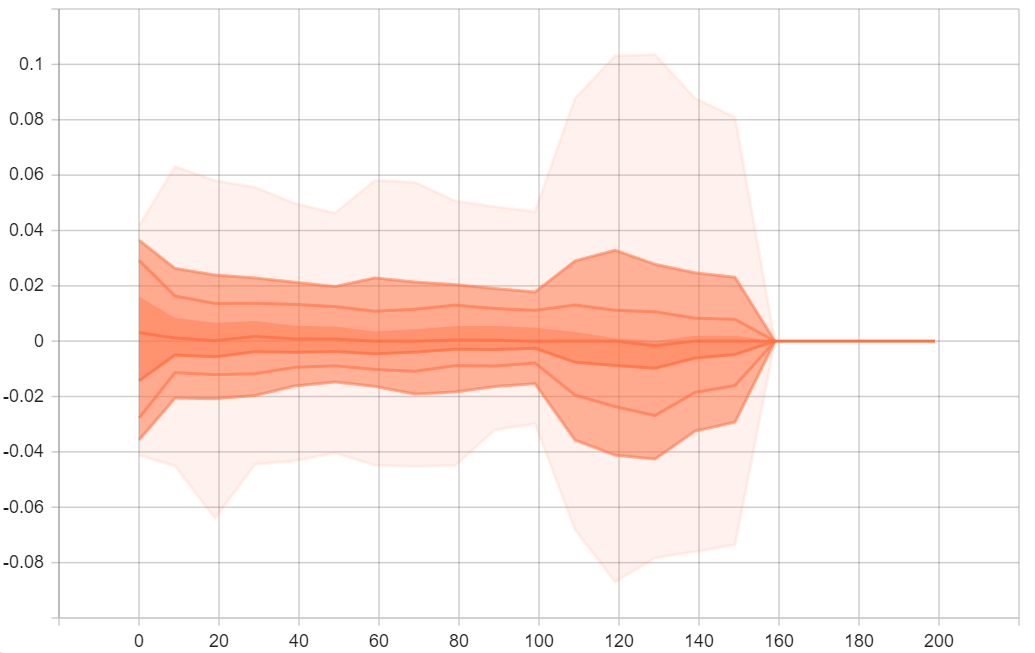} }}%
    \qquad
    \subfloat[\centering Layer 12 Output channel 50 ]{{\includegraphics[width=2cm]{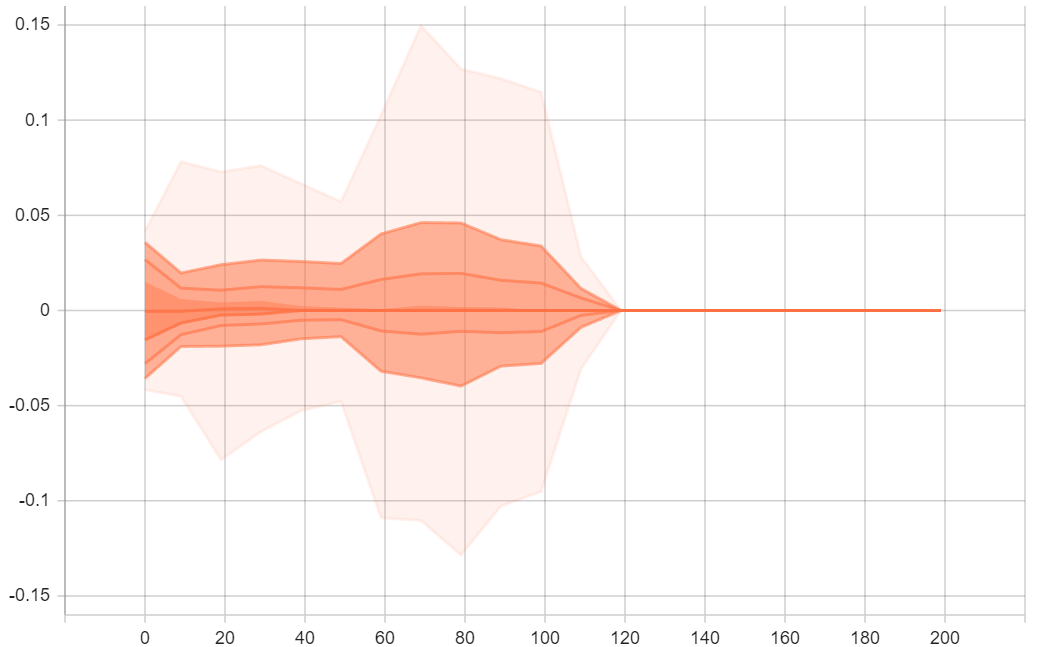} }}%
    \qquad
    \subfloat[\centering Layer 16 Output channel 100]{{\includegraphics[width=2cm]{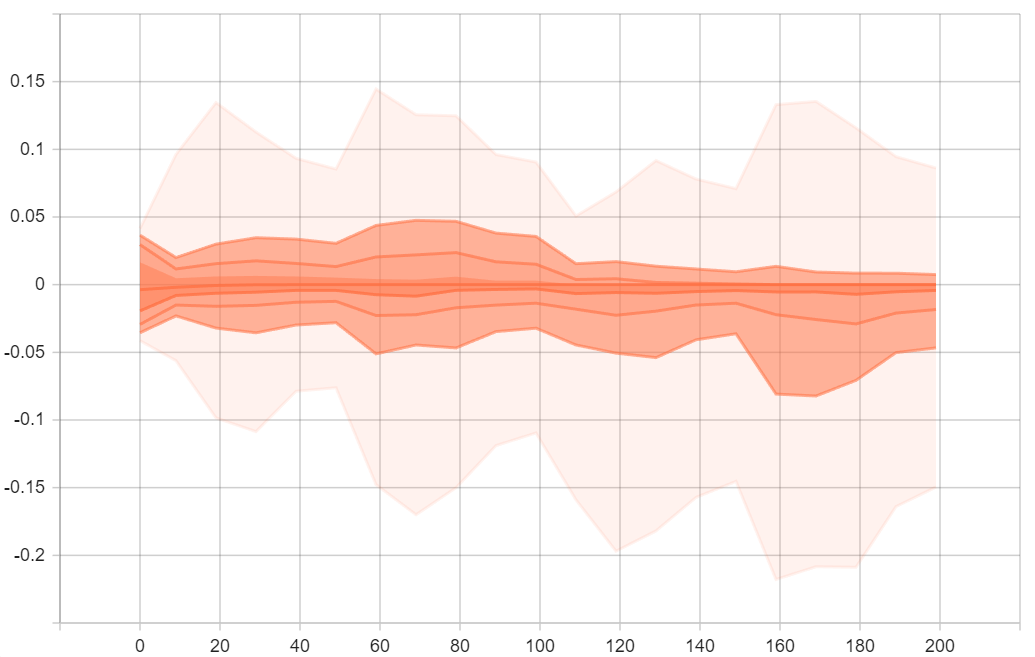} }}%
    \qquad
    \subfloat[\centering Cyclical learning rate]{{\includegraphics[width=2cm]{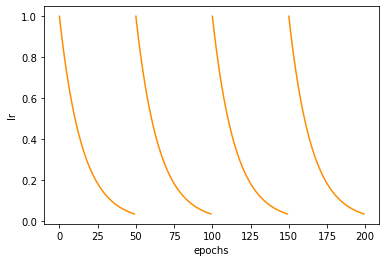} }}%
    \caption{Sample paths of all weights from select channels of ResNet18 in CIFAR10 during training with cyclical learning rate $l_{t}$ and $k=\frac{1}{N}$ from algorithm 2. (a)-(e) are grouped by either input and output channels, (f) is the cyclical learning rate used.}%
    \label{fig:1}
\end{figure}

However, \cite{gur2018gradient} argued that only a tiny subspace of the parameters have non-zero gradients. Empirically, when we run algorithm 2, we find that the gradients will vanish everywhere except in a small  number of input and output channels. In addition, during optimization, many of parameters in the input and out channels will concentrate to zero. These concentration is quite stable even under the two following perturbations from the algorithm:\\
1. A cyclical learning rate, which can jump from small to large. See Figure \ref{fig:1}-(f). \\
2. At each iteration, adding an injected noise $\sqrt{2+2\beta_{1}^{2}}k(\Delta t)^{1.5}\boldsymbol{\epsilon}$.

\begin{figure}[t]
    \centering
    \includegraphics[width=2.6cm]{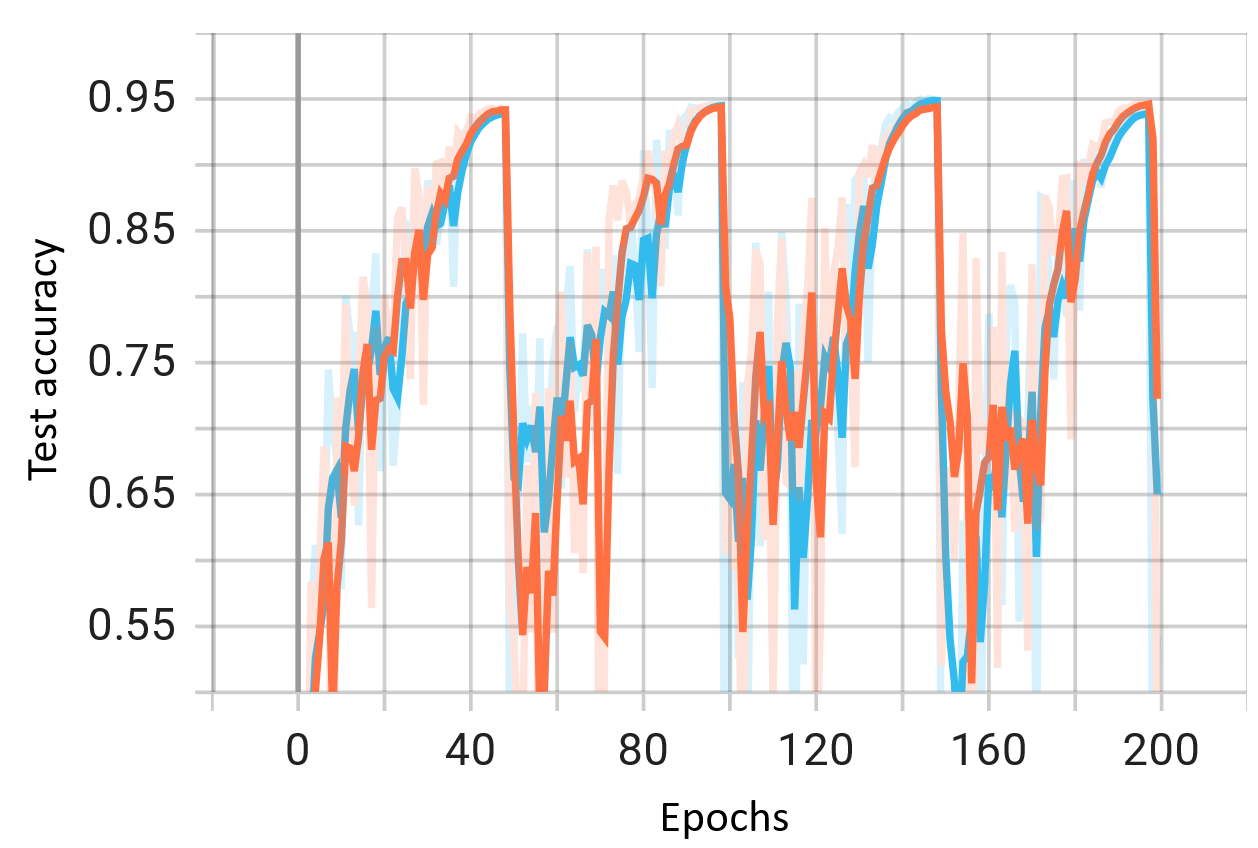}
    \includegraphics[width=2.6cm]{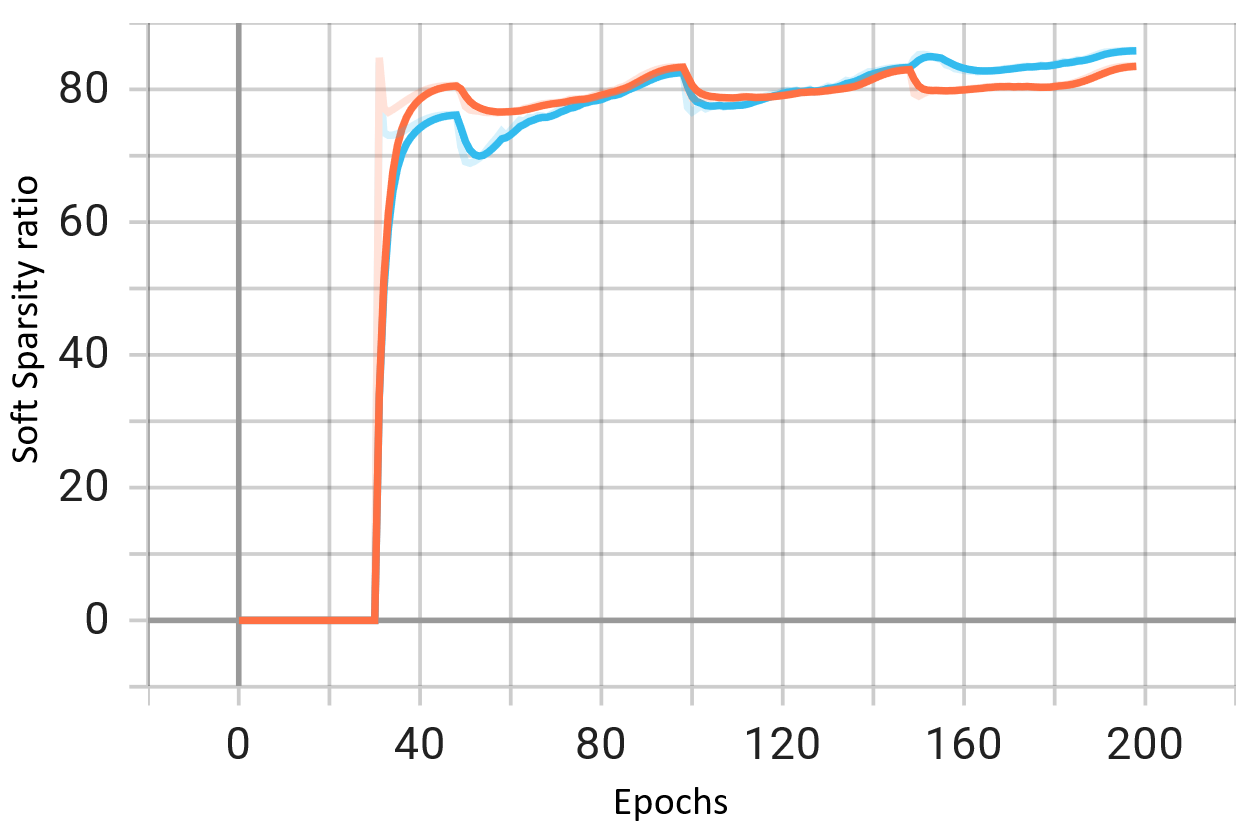}
    \includegraphics[width=2.6cm]{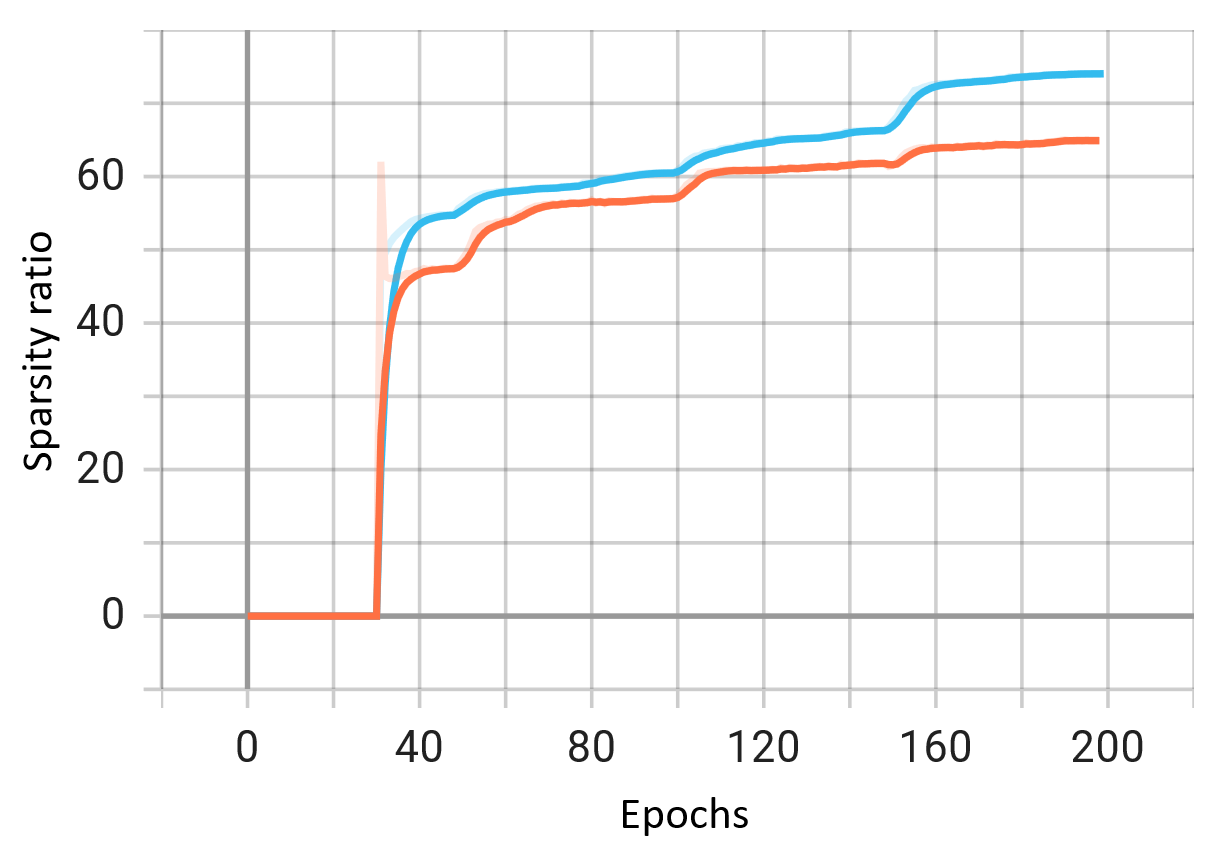}
    \caption{Resnet18 pruned using algorithm 3 with the cyclical learning rate under DFP  (blue) and DPF (orange), under the same hyper-parameter setting. The left panel shows testing accuracy. The middle panel shows the percentage of the weights, which use large weight decay factor as suggested by EM algorithm (soft sparsity ratio). The right panel shows the sparsity ratio of the model during training.}
    \label{fig:2}
\end{figure}

We observe the concentration phenomenon by visualizing the dynamic distributions of the weights in Figure \ref{fig:1}, which shows the sample paths of all weights from select channels during optimization. Note that, at this stage, we just ran our algorithm 2 with cyclical learning rate. 
We can see that the contractions occur regardless of whether the grouping is based on input or output channels. In addition, we rarely observe regrowth of the weight once the whole channel concentrates to zero, see Figure \ref{fig:2}. This phenomenon is very similar to the posterior contraction in high dimensional sparse Bayesian linear regression \citep{castillo2015bayesian}, where only a sub-model has substantial probability mass. Finally, Figure \ref{fig:1}$(e)$ shows the  behavior of weights in those channels which are not concentrated around zero.

Motivated by the above observations,  we propose a structured pruning rule based on a simple concentration metric, where, within each layer, two for loops will be used to scan through all the input channels and output channels. 
\begin{equation}\label{pruning_criterion1}
\begin{aligned}
\text{If} & \,\, \max_{j,k} |W_{ijk}|-\min_{j,k} |W_{ijk}|<\lambda_{2}  \,\, \text{set} \,\, W_{i}=0;\\
\text{If} & \,\, \max_{i,k} |W_{ijk}|-\min_{i,k} |W_{ijk}|<\lambda_{2} \,\, \text{set} \,\, W_{j}=0;
\end{aligned}
\end{equation}
where $\lambda_{2}>0$ is a hard threshold which we need to tune.  Algorithm 3 provides the pseudo code. The SoftMask returned by the EM algorithm will control the shrinkage of the weight. It is used to separate the noise and signal as in sparse linear regression \citep{castillo2015bayesian}. The group of weights with very small magnitude will be identified as noise and a strong shrinkage (a large weight decay) will be applied to it in the next iteration and vice versa. The HardMask returned by the pruning criterion rule will remove the weight. Although we recommend using DFP, where regrowth is not allowed, we also provide the user with DPF as another option in Algorithm 3. Figure \ref{fig:2} shows no significant differences between the DFP and DPF, when comparing their test accuracy and sparsity ratios. \\
\textbf{Remark}: The pruning criterion from equation (\ref{pruning_criterion1}) assumes that if the weights within the input channels or output channels concentrate together, it will concentrate to zero. However, the strong concentration phenomenon we observed in CNN as shown in Figure $\ref{fig:1}$ are observed in feedforward neural networks, we find that even when the problem is known to be sparse, this criterion can lead to a lack of convergence. Instead, we propose the following $L_{2}$ magnitude pruning criterion 
\begin{equation}\label{L2}
\text{If} \,\,\sum_{i=1}^{k_{o}}w_{ij}^{2}/k_{o} \leq \lambda_{1},\text{set}\,\, w_{j}=0    
\end{equation}
where $i$ is output units, $j$ is input units and $k_{o}$ is the number of output units.

\begin{algorithm}
\centering
\caption{EM-MCMC}
\begin{algorithmic}
\State \textbf{Input} $\beta=(\beta_{1},\beta_2)$, $(\delta_{0},\delta_{1}) \text{(weight decay)}$\\
\quad \quad \,\,$N\text{(training size)}$, $B\text{(batch size)}$\\
\quad \quad \,\,$\lambda_{1}\text{(soft threshold)}$, $\lambda_{2}\text{(hard threshold)}$\\
\quad \quad \,\,$\text{pruning with feedback}=\text{False}$
\State\textbf{Initialization}  $\small{\boldsymbol{\rho},\boldsymbol{w},\boldsymbol{v},\text{SoftMask} \gets 1,\text{HardMask} \gets 1}$
\For{$t=1 \: \cdots \: $}
\State $\delta \gets \delta_{1}$
\State $\delta[\text{SoftMask}==0] \gets \delta_{0}$
\State $\Delta t \gets s_{1}(t)$; $\eta \gets s_{2}(t)$; $k \gets T(t)$;
\State $\alpha \gets k*\Delta t^{1.5}$
\State $\boldsymbol{g} \gets -\frac{1}{B}\sum_{i \in \mathcal{B}}\nabla_{\mathbf{w}}\log(\mathcal{D}_{i}|\mathbf{w})+\frac{\delta}{N} \mathbf{w}$
\State $\boldsymbol{\rho}  \gets   \boldsymbol{\rho}+ \left(\frac{\eta}{\boldsymbol{\tau}}-\eta\alpha^{2}\delta \boldsymbol{\tau}\right)- \eta\alpha\boldsymbol{\epsilon} \odot \boldsymbol{g}$
\State $\boldsymbol{\tau} \gets   1/(1+e^{-\boldsymbol{\rho}})$
\State $\boldsymbol{\epsilon} \gets N(0,\mathbf{1}_{p})$
\State $\small{\boldsymbol{v} \gets\beta_{1}\boldsymbol{v}-\beta_{2}g(\Delta t)^{2}+\sqrt{2+2\beta_{1}^{2}}k(\Delta t)^{1.5}\epsilon}$
\State $\mathbf{w} \gets \mathbf{w} - \tau \odot \boldsymbol{v}$
\If{$t>\text{warm up}$}
\State $\text{SoftMask} \gets \textbf{EM}(\mathbf{w},\lambda_{1})$
\State $\text{HardMask} \gets \textbf{Pruning Criterion}(\mathbf{w}, \lambda_{2})$
\State $\mathbf{w}[\text{HardMask}] \gets 0$
\If{$\text{pruning with feedback is False}$}
\State $\text{Freeze}\,\, \mathbf{w}[\text{HardMask}]$
\EndIf
\EndIf
\EndFor
\end{algorithmic}
\end{algorithm}

\section{Experiments}
\subsection{Experimental setup for image classfication task}
In this section, we conduct experiments to show the effectiveness of our proposed algorithms in terms of the test accuracy and sparsity ratio by comparing against both dense and sparse methods in the literature for image classfication. Apart from the baseline, which is SGD with momentum, we include two SOAT Bayesian methods: Rank-1 BNN \citep{dusenberry2020efficient}, a variational inference approach with batch ensemble \citep{wen2020batchensemble}; and an MCMC approach cSGLD \citep{zhang2019cyclical}. We also include three popular dynamic sparse training methods: SM \citep{dettmers2019sparse}; DSR\citep{mostafa2019parameter} and DPF \citep{lin2020dynamic}. \textbf{We repeat each experiment three times and average them to obtain stable results}. 

The Rank-1 BNN has some advantage in this experimental comparison as their training approach is more computationally intensive. Since we don't have enough TPU to implement their method, we extracted their experimental result from the original paper as indicated by * in the table. 

For the other methods, we use 200 epochs to train models in CIFAR10 and CIFAR100 datasets with single RTX3090 node and 200 epochs to train model in ImageNet dataset with single A100 node. The cyclical learning rate with 4 cycles as shown in Figure \ref{fig:1}-(f) is used for cSGLD \citep{zhang2019cyclical} and CV-Adam. We collect three samples at the end of each cycle,  which gives us 12 samples in total for CIFAR10 for CIFAR100 datasets, while for ImageNet, the first 50 epochs is used as warm up, therefore, we only collect 9 samples. The posterior predictive distribution is calculated by averaging all the samples as an ensemble.

\textbf{Since DPF equation (\ref{DPF}) has no official code release, but easy to insert into Algorithm 3, we implement it by ourselves}. For the baseline model, we follow the hyper-parameter settings from \cite{xie2022adaptive}. Their setting allows us to train a strong baseline for comparison. Details of hyper-parameter settings is provided in Appendix D.

\subsection{Results for image classfication task}

\begin{table*}[!]
\centering
\caption{Comparison of test accuracy with different target sparsity ratios in CIFAR10.}
\begin{tabular}{ccccccccccc} 
\hline
          & \multicolumn{4}{c}{Dense Method}         &  & \multicolumn{5}{c}{Sparse Method}                   \\ 
\cline{2-5}\cline{7-11}
Model     & Baseline & Rank-1 BNN & cSGLD  & CV-Adam &  & SM     & DSR    & DFP    & DPF    & Sparsity ratio  \\ 
\hline
ResNet18  & 95\%     & -          & 95.7\% & 95.5\%  &  & 92.8\% & 93.1\% & 94.7\% & 94.8\% & 70\%            \\
ResNet18  & 95\%     & -          & 95.7\% & 95.5\%  &  & 91.1\% & 91.2\% & 94.5\% & 94.5\% & 80\%            \\
ResNet18  & 95\%     & -          & 95.7\% & 95.5\%  &  & 89.7\% & 90.0\% & 93.9\% & 93.8\% & 90\%            \\ 
\hline
WRN-28-10 & 96.3\%   & 96.5\%*    & 96.2\% & 96.5\%  &  & 95.4\% & 95.8\% & 95.5\% & 95.8\% & 90\%            \\
WRN-28-10 & 96.3\%   & 96.5\%*    & 96.2\% & 96.5\%  &  & 95.3\% & 95.4\% & 95.3\% & 95.4\% & 95\%            \\
WRN-28-10 & 96.3\%   & 96.5\%*    & 96.2\% & 96.5\%  &  & 94.5\% & 94.5\% & 94.7\% & 94.5\% & 99\%            \\
\hline
\end{tabular}
\end{table*}
\begin{table*}[!]
\centering
\caption{Comparison of test accuracy for different target sparsity ratios for in CIFAR100}
\begin{tabular}{ccccccccccc} 
\hline
          & \multicolumn{4}{c}{Dense method}         &  & \multicolumn{5}{c}{Sparse method}                   \\ 
\cline{2-5}\cline{7-11}
Model     & Baseline & Rank-1 BNN & cSGLM  & CV-Adam &  & SM     & DSR    & DFP    & DPF    & Sparsity ratio  \\ 
\hline
ResNet34  & 78.5\%   & -          & 79.7\% & 79.4\%  &  & 76.4\% & 76.6\% & 77.6\% & 77.4\% & 50\%            \\
ResNet34  & 78.5\%   & -          & 79.7\% & 79.4\%  &  & 73.2\% & 73.8\% & 75.4\% & 75.1\% & 70\%            \\
ResNet34  & 78.5\%   & -          & 79.7\% & 79.4\%  &  & 72.1\% & 72.3\% & 75.1\% & 75.3\% & 90\%            \\ 
\hline
WRN-28-10 & 81.7\%   & 82.4\%*    & 82.7\% & 83.4\%  &  & 78.0\% & 78.1\% & 78.5\% & 78.2\% & 70\%            \\
WRN-28-10 & 81.7\%   & 82.4\%*    & 82.7\% & 83.4\%  &  & 77.0\% & 76.5\% & 77.7\% & 77.9\% & 80\%            \\
WRN-28-10 & 81.7\%   & 82.4\%*    & 82.7\% & 83.4\%  &  & 76.5\% & 76.4\% & 77.5\% & 77.4\% & 90\%            \\
\hline
\end{tabular}
\end{table*}
\begin{table*}[!]
\centering
\caption{Comparison of test accuracy for different target sparsity ratios for ImageNet}
\begin{tabular}{ccccccccccc} 
\hline
         & \multicolumn{4}{c}{Dense method}         &  & \multicolumn{5}{c}{Sparse method}                   \\ 
\cline{2-5}\cline{7-11}
Model    & Baseline & Rank-1 BNN & cSGLM  & CV-Adam &  & SM     & DSR    & DFP    & DPF    & Sparsity ratio  \\ 
\hline
ResRet50 & 76.5\%   & 77.3\%*    & 76.7\% & 76.8\%  &  & 73.8\% & 73.3\% & 74.5\% & 74.6\% & 80\%            \\
ResNet50 & 76.5\%   & 77.3\%*    & 76.7\% & 76.8\%  &  & 72.3\% & 72.0\% & 73.5\% & 73.2\% & 90\%            \\
\hline
\end{tabular}
\end{table*}

\begin{table*}[h]
\centering
\caption{Comparison of sparse methods for three simulated examples}
\begin{tabular}{ccccccccccccccc} 
\cline{1-5}\cline{7-10}\cline{12-15}
       & \multicolumn{4}{c}{Example 1}  &  & \multicolumn{4}{c}{Example 2} &  & \multicolumn{4}{c}{Example 3}   \\ 
\cline{1-5}\cline{7-10}\cline{12-15}
Method & MSE  & FDR  & FNDR & $\hat{S}$ &  & MSE  & FDR  & FNDR & $\hat{S}$        &  & Accuracy & FDR    & FNDR & $\hat{S}$    \\ 
\cline{1-5}\cline{7-10}\cline{12-15}
SPLBNN & 1.4  & 0    & 0    & 4         &  & 2.43 & 0    & 0    & 5        &  & 91.2\%   & 0      & 0    & 5    \\
SVBNN  & 1.6  & 38.3\% & 0    & 6.3       &  & 3.12 & 51\% & 0    & 10.2     &  & 86\%     & 39.9\% & 0    & 8.2  \\
DFP    & 1.32 & 0    & 0    & 4         &  & 2.46 & 0    & 0    & 5        &  & 94.70\%  & 0      & 0    & 5    \\
DPF    & 1.29 & 0    & 0    & 4         &  & 2.49 & 0    & 0    & 5        &  & 94.50\%  & 0      & 0    & 5    \\
\cline{1-5}\cline{7-10}\cline{12-15}
\end{tabular}
\end{table*}
\textbf{Performance for CIFAR10 and CIFFAR100 dataset}:
We report the testing accuracy and sparsity ratio for sparse method in Table 1 and Table 2. For the dense method, comparing with two SOTA Bayesian methods and baseline, the CV-Adam of Algorithm 1 is very competitive in both CIFAR10 and CIFAR100 datasets. 

When pruning is applied, there is almost no performance loss with both DFP and DPF approaches compared with the dense methods in ResNet18,  and a much higher sparsity ratio can be achieved by WRN-28-10 in the CIFAR10 dataset. 
For CIFAR100 dataset, we found that it is harder to target the high sparsity ratio while at the same time keep the performance loss negligible. 
Overall, our pruning algorithms can achieve high sparsity ratio while sacrificing a small amount in accuracy. 

Finally,  there is no evidence that using a dynamic forward pruning with (no regrowth suffers performance loss compared to DPF.

\textbf{Performance for ImageNet}: As shown in Table 3, 
CV-Adam produced a modest performance gain compared with the baseline, but performed worse than Rank-1 BNN.
For cSGLM\citep{zhang2019cyclical}, we failed to reproduce the $77\%$ predictive accuracy as claim in the original paper. We believe the performance for both our CV-Adam and cSGLM can be further improved by tuning the hyper-parameters carefully. For sparse method, our pruning algorithms out performed the other two methods SM and DSR. Again, there is still no evidence that using DFP will incur performance loss.

\subsection{Simulated Examples: variable selection for nonlinear regression}
\cite{sun2021consistent} applied the Spike-and-Slab Gaussian
prior to prune the neural network (SPLBNN). They use a Laplace approximation-based approach to approximate the marginal posterior inclusion probability, which requires the user to train the dense model to find the local mode first followed by pruning and retraining the sparse model, finally they use the Bayesian evidence to elicit sparse DNNs in multiple runs with different initialization. Another related approach is variational BNN with Spike-and-Slab prior \citep{bai2020efficient} (SVBNN). Both approaches are not scalable to large models, but they work well in high dimensional sparse nonlinear regression, so we carry out a comparison with these two approaches in this setting. Here, we follow the same data generate process and examples as described in \cite{sun2021consistent}:
\begin{itemize}
\item  Simulate $e,z_{1},...,z_{p}$ independently from the truncated standard normal distribution on the interval $[-10,10]$.
\item Set $x_{i}=\frac{e+z_{i}}{\sqrt{2}}$ for $i=1,...,p$
\end{itemize}
Then, all the predictor $x_{i}$ fall into a compact set and are mutually correlated with a correlation coefficient of about 0.5.
Based on this setting, we generate three toy examples from \cite{sun2021consistent} where each example consists of 10000 training samples and 1000 testing samples. To fit the data and make comparison, we follow the network structure described in \cite{sun2021consistent}.\\
\textbf{Example 1}:
$$
\begin{aligned}
y=& \tanh \left(2 \tanh \left(2 x_1-x_2\right)\right)+2 \tanh \left(\tanh \left(x_3-2 x_4\right)\right.\\
&\left.-\tanh \left(2 x_5\right)\right)++0 x_6+\cdots+0 x_{1000}+\varepsilon,
\end{aligned}
$$
\textbf{Network Structure}: 1000-5-3-1 with ReLu activation\\\\
\textbf{Example 2}:
$$
y=\frac{5 x_2}{1+x_1^2}+5 \sin \left(x_3 x_4\right)+2 x_5+0 x_6+\cdots+0 x_{2000}+\varepsilon
$$
\textbf{Network Structure}: 2000-6-4-3-1 with ReLu activation\\\\
\textbf{Example 3}:
$$
y= \begin{cases}1 & e^{x_1}+x_2^2+5 \sin \left(x_3 x_4\right)+0 x_5+\cdots+0 x_{1000}>3\\ 
0 & \text { otherwise. }\end{cases}
$$
\textbf{Network Structure}: 2000-6-4-3-1 with ReLu activation\\\\
where $\epsilon \sim N(0,1)$. We only compare our algorithm with the method from  \cite{sun2021consistent} and \citep{bai2020efficient}. The other two sparse methods we used in the image task didn't work in these three examples. We use equation (\ref{L2}) as the pruning criterion in Algorithm 3. The performance of variable selection is measured by the false discover rate (FDR) and false non-discover rate (FNDR). The predictive mean square error is used for examples 1 and 2 and the prediction accuracy is used for example 3. $\hat{S}$ is the number of the selected variables returned by the competing methods. We repeat the experiments 10 times and report the averaged results in Table 4.  Details of hyper-parameter setting is provided in the Appendix D.

\textbf{Performance analysis}: In terms of variable selection, SPLBNN, DFP and DPF perfectly selected all the relevant variables and removed all irrelevant variables in all three examples.  SVBNN always selected 
more variables but was able to  keep the FNDR at zero. The predictive performance for examples 1 and 3 are good for SPLBNN, DFP and DPF. But for example 2, the mean square error are high (compared to the oracle value of 1), A better network structure may improve this performance. One of the surprising results is that DFP can still identify all relevant predictors and was never trapped in a bad local mode. Comparing with the SPLBNN from \cite{sun2021consistent}, which requires the train-prune-fine tune process multiple times, our algorithm is very aggressive and efficient and it outperforms the SPLBNN in example 1 and 3.

\section{Discussion of future work}
In this paper, we showed the connection between our variational BNN algorithm with stochastic gradient Hamiltonian Monte Carlo. The cyclical learning rate has been used with the hope that we can jump to different modes by suddenly changing the learning rate from small to large. Using the posterior predictive distributions in an ensemble approach can improve the predictive accuracy. One question arises after observing the concentration phenomenon from CNN is do we really need to train the full model after each cycle? Our experimental findings suggest that only a small number of the subnetworks needs to be trained after one cycle. Maybe the mode jump 
is due to the change of a small proportion of the weights from subnetworks. If this is true, it implies that training can be accelerated by developing the dynamic forward freeze algorithm to gradually freeze the parameters at the end of each cycle.
In addition, we can use the subnetwork ensemble approach where only small proportion of the weights are different among different models and majority of the weights are shared cross all the models. This approach can save storage requirements and speed up computation.

\bibliographystyle{apalike}
\bibliography{main}

\end{document}